\documentclass[letterpaper]{article} % DO NOT CHANGE THIS
\usepackage{aaai2027}  % DO NOT CHANGE THIS
\nocopyright

\usepackage[hyphens]{url}  % DO NOT CHANGE THIS
\usepackage{graphicx} % DO NOT CHANGE THIS
\usepackage{natbib}  % DO NOT CHANGE THIS AND DO NOT ADD ANY OPTIONS TO IT
\usepackage{caption} % DO NOT CHANGE THIS AND DO NOT ADD ANY OPTIONS TO IT
\usepackage{algorithm}
\usepackage{algorithmic}
\usepackage{amsmath}
\usepackage{amsfonts}
\usepackage{comment}
\usepackage{xr}
\usepackage{newfloat}
\usepackage{listings}
\DeclareCaptionStyle{ruled}{labelfont=normalfont,labelsep=colon,strut=off} % DO NOT CHANGE THIS
\floatstyle{ruled}
\newfloat{listing}{tb}{lst}{}
\floatname{listing}{Listing}

\usepackage{booktabs}
\usepackage{makecell}
\usepackage{array} % needed for centered, fixed-width columns in the author table
\title{GUIDE: Generative Utility Inference and Decision Engine}
\title{GUIDE: Generative Utility Inference and Decision Engine}

\author{
\centering
\begin{tabular}{
>{\centering\arraybackslash}p{0.31\linewidth}
>{\centering\arraybackslash}p{0.31\linewidth}
>{\centering\arraybackslash}p{0.31\linewidth}}
\textbf{Anagha Tiwari} &
\textbf{Alexander G. Gray} &
\textbf{Nick Feamster} \\[3pt]
\normalfont\small Department of Computer Science &
\normalfont\small &
\normalfont\small Department of Computer Science \\
\small University of Chicago &
\small\vspace{-20pt} Centaur AI Institute &
\small University of Chicago \\
\small\path{anaghatiwari@uchicago.edu} &
\small\vspace{-19pt}\path{alexander.gray@centaurinstitute.org} &
\small\path{feamster@uchicago.edu} \\[12pt]
\textbf{Brian Jabarian} &
\textbf{Alex Imas} &
\textbf{Alex Kale} \\[3pt]
\normalfont\small Heinz College \& School of Computer Science\footnote{Heinz College of Information Systems and Public Policy and, by courtesy, the Human-Computer Interaction Institute, School of Computer Science, Carnegie Mellon University} &
\normalfont\small Booth School of Business &
\normalfont\small Department of Computer Science \\
\small Carnegie Mellon University &
\small\vspace{-15pt} University of Chicago &
\small\vspace{-15pt} University of Chicago \\
\small\path{jabarian@cmu.edu} &
\small\path{alex.oleg.imas@gmail.com} &
\small\path{kalea@uchicago.edu}
\end{tabular}
}

\begin{document}
\affiliations{}
\maketitle

\begin{abstract}

% Utility elicitation and alignment remain fundamental challenges for AI systems. 
Measuring the preferences of human users remains a fundamental challenge of AI alignment. Existing elicitation approaches struggle to efficiently discover multidimensional preferences or accurately ground these inferences in domain knowledge. To address this, we introduce GUIDE, an LLM-driven elicitation architecture that infers user preferences through conversations by combining Bayesian adaptive sampling for question selection and symbolic representation learning to initialize domain-specific preference models. GUIDE generalizes adaptive sampling to diverse elicitation questions through an extensible type system of transforms on a parameterized preference state. GUIDE produces domain-specific preference representations through an initialization process using symbolic rule-based learning to capture world knowledge and set priors over preference dimensions grounded in data about decision alternatives. The architecture provides observability and steerability to facilitate deployment and analyze elicitation processes. In silico experiments on investment portfolio optimization demonstrate that GUIDE improves cold-start and minimizes recommendation regret consistently within early elicitation interactions across user personas compared to prior work, LLM-only baselines, and ablated GUIDE versions.

% elicitnig complex pref from heter users in way that can be sacled across domains

\end{abstract}

\section{Introduction}

% , experiences, and values for informed decision making on different levels (\citet{pe-healthcare}; \citet{pe-risk}; \citet{pe-finance})

% Preference elicitation (PE), a method for measuring preferences and desired outcomes, is  essential for aligning AI systems and is applied to support high-stakes decisions in a variety of settings (e.g., \citealp{pe-healthcare}; \citealp{pe-risk}; \citealp{pe-finance})). 
Preference elicitation (PE) -- the process of measuring and modeling human preferences and desired outcomes -- is fundamental to both informed decision-making and AI alignment.
% 1) informed decision-making, and 2) AI alignment, where systems must accurately infer and represent human values and intentions to make decisions.
% Applications in healthcare, consumer services, finance, and more use PE to personalize recommendations and support high-stakes decisions (e.g., \citealp{pe-healthcare}; \citealp{pe-risk}; \citealp{pe-finance}). 
In finance, for example, PE is used to capture investors' risk preferences, construct personalized portfolios, manage financial exposure, and support compliance with regulatory standards \cite{CHEN2026102565}. However, failing to accurately capture user preferences can introduce bias, misrepresent population characteristics, distort estimates of users' true preferences, and ultimately degrade downstream decision quality \cite{10.1145/3437963.3441786}. Especially in settings where user preferences are complex and heterogeneous, PE approaches must be capable of reliable inference from nuanced human inputs~\cite{Cuthbertson_Penney_2023, pol-decision-making}.

% Bayesian methods provide an initial solution 
Existing Bayesian approaches have improved PE, but they remain constrained by the preference information they can acquire. By maintaining uncertainty over a user's latent utility function, selecting queries, and updating the posterior with new evidence, Bayesian methods can efficiently identify informative questions \cite{CHAPMAN202525}. However, most current approaches use two-alternative forced choice 
% or binary 
questions, which are structurally limited in terms of the information gain in a single elicitation turn (\citealp{10.1145/3450613.3456814};
\citealp{vendrov2019gradientbasedoptimizationbayesianpreference};
\citealp{10.1145/2939672.2939746}; \citealp{pmlr-v9-guo10b};
\citealp{neiswanger2022generalizingbayesianoptimizationdecisiontheoretic}). As a result, these approaches require many elicitation cycles drawn from a narrow range of queries, and struggle to capture complex preferences more naturally expressed through open-ended, unstructured elicitation formats.

% In this sense, current Bayesian methods do not effectively leverage the generative capacity of LLMs as question generators.

% Although LLMs enable natural-language dialogue and question generation (\citealp{li2023elicitinghumanpreferenceslanguage}; \citealp{montazeralghaem2025askingclarifyingquestionspreference}), they lack an explicitly structured representation of user preferences, uncertainty, and multi-step elicitation processes (\cite{capstick2025autoelicitusinglargelanguage}).
Recent work has therefore used Large Language Models (LLMs) directly for PE, as they allow users to express preferences through natural-language dialogue and can generate informative elicitation questions (\citealp{li2023elicitinghumanpreferenceslanguage}; \citealp{montazeralghaem2025askingclarifyingquestionspreference}). However, LLMs used as standalone PE systems face two related limitations.
% However, when used as standalone PE systems, 
First, they lack an explicit representation of preference uncertainty and a principled mechanism for reasoning consistently across multi-step elicitation (\cite{capstick2025autoelicitusinglargelanguage}). 
% As a result,
Second, because their domain knowledge is implicit rather than explicitly modeled, LLMs can generate inconsistent inferences, fixate on irrelevant conversational details, hallucinate, and ultimately produce recommendations poorly grounded in domain-specific knowledge or user preferences (\citealp{Ferrara_2023}; \citealp{yang2023foundationmodelsdecisionmaking}). 
% Second, the implicit domain models embedded in LLMs may generate inconsistent inferences, fixate on irrelevant conversational details, hallucinate, and ultimately produce recommendations that are poorly grounded in domain-specific knowledge or user preferences (\citealp{Ferrara_2023}; \citealp{yang2023foundationmodelsdecisionmaking}). 
These limitations motivate using LLMs as interfaces within a structured elicitation process rather than as the preference model itself. They also suggest the need for human oversight and steering in order to align LLM-based PE procedures with domain-specific knowledge.

% Building on the above limitations of existing methods and contributions with potential techniques, 
Given these challenges, we argue that a PE system capable of supporting diverse decision problems must satisfy three requirements: (R1) \textit{expressive querying}: the ability to efficiently ask diverse question types to users and reliably interpret their responses; (R2) \textit{empirical grounding}: grounding inferences in empirical domain knowledge that constrains the space of plausible preferences and outcomes; and (R3) \textit{observability}: an elicitation process transparent enough to enable expert oversight, drive systematic analysis, and facilitate adaptation across domains.

% To satisfy these requirements, we
We introduce GUIDE (Generative Utility Inference and Decision Engine), an LLM-assisted PE framework 
that 
% enables Bayesian adaptive sampling with
meets all three requirements within a single architecture. \textbf{To this end, we contribute:} (i) a Bayesian elicitation framework featuring dimension discovery and preference calibration processes driven by a diverse, extensible set of question types~(R1), (ii) a symbolic domain initialization pipeline to construct interpretable rules and population priors for domain specific grounding~(R2), and (iii) a transparent structure that enables analysis of elicitation interactions and outputs, model coverage, and expert steering and oversight~(R3).

    % These contributions are 1) validated via a \textbf{quantitative evaluation} of GUIDE 
We evaluate GUIDE in the domain of financial portfolio optimization to (1) validate its ability to recover ground-truth preferences across a diverse set of investor profiles and (2) compare its performance against ablated variants, LLM-only approaches, and prior work, specifically OPEN and PEBOL (\citealp{handa2024bayesianpreferenceelicitationlanguage}, \citealp{Austin_2024}). Across these evaluations, we show that GUIDE consistently improves cold-start and early-turn recommendation quality under limited interaction budgets, while heterogeneous transforms introduce a trade-off in late-stage stability.

% Beyond the performance these components deliver, w
% Since the engine's preference representation is explicit with editable and observable domain knowledge and elicitation question updates, 
We regard GUIDE's wider scientific contribution as making the design of LLM-based PE methods more learnable. Since the engine's preference representation and policies for generating and selecting elicitation questions are explicit, observable, and editable, our architecture turns design choices that are usually entangled inside a prompt or a model's weights into components that can be studied systematically. We therefore present GUIDE as a framework for studying elicitation strategies, adaptive question selection, and interactions between statistical and generative AI methods, while enabling principled research on elicitation dynamics.

\section{Preliminaries and Related Works}

We aim to accurately elicit and represent user preferences via an adaptive sampling approach with LLM-assisted question generation.  Our approach adapts and extends the OPEN framework, which uses Bayesian Optimal Experimental Design to select elicitation questions in order to learn a posterior over preference states (\citet{handa2024bayesianpreferenceelicitationlanguage}). The core mechanism we adapt is next-question selection based on expected information gain (EIG) in a particle-filter utility model.
% We use previous approaches on elicitation that leverage Bayesian models to maintain and update a posterior over preference states, specifically the OPEN framework that utilizes Bayesian Optimal Experimental Design with LLM-assisted feature extraction and query translation \citet{handa2024bayesianpreferenceelicitationlanguage}. 

However, whereas OPEN is limited to pairwise comparisons over combinations of decision alternatives, GUIDE supports a diverse, extensible set of question types (e.g., pairwise, rule elicitation, dimension dominance, dimension proposal, and free-text inputs) and corresponding transforms to the underlying preference state. Further, since OPEN relies on LLMs to confabulate plausible features of decision alternatives based on LLM-generated domain descriptions, its parameterizations of users' utility models may not actually be predictive in real-world decision problems. To address this problem, GUIDE's preference dimensions are learned from and thus empirically grounded in real data about decision alternatives in the domain through an initialization process.
% However, OPEN is limited to using pairwise question transforms for elicitation, therefore constraining the information that can be gathered in each interaction cycle. We enable GUIDE to utilize diverse question transforms (including pairwise and rule elicitation, dimension dominance, dimension proposal, and free-text inputs) to elicit the most relevant information from a user, where next-question selected is done via an expected information gain (EIG) competition among transform types. 

\textbf{Decision space.} There are $K$ decision alternatives. These are domain-specific recommendation options with measured feature values, either provided by a developer or generated from previous data. For example, in portfolio management, model portfolios scored on risk level, volatility, returns, fees, etc. Each alternative is an $F$-dimensional continuous feature vector $\mathbf{f}_k \in [0,1]^F$ normalized to the unit interval.

\textbf{Dimensions and loading mappings.} A preference dimension $d$ is a latent evaluative axis (e.g., ``risk tolerance'') defined by signed loadings over portfolio features, where the sign indicates preference direction and the magnitude indicates feature importance. GUIDE supports both developer-defined starting axes and LLM-proposed axes discovered mid-dialogue, which are validated before use. Nonlinear or interior preferences are represented through features that encode the desired shape (e.g., a diversification-balance feature peaking at a 50/50 equity/bond split). Definitions and implementation are in Supplementals section A.1.

\textbf{Elicitation questions as transform types.} 
A transform maps a user's response to a question into evidence in the posterior.
% into evidence on the belief. 
Formally, each transform type defines, for a candidate question $q$, a likelihood $p(y\mid \mathbf{w},q)$ over its possible answers $y$; observing $y$ then updates every particle's log posterior weight additively, $\ell^{(i)} \mathrel{+}= \log p(y\mid \mathbf{w}^{(i)},q)$, while leaving particle locations fixed. Different transform types use different likelihoods but share a single update form, allowing heterogeneous questions to act on a single common posterior.

\textbf{Utility and posterior.} Utility is linear in the dimension scores, $u_k(\mathbf{w}) = \sum_d \tilde w_d\,\phi_d(\mathbf{f}_k)$, with a non-negative dimension-weight vector $\mathbf{w}\in\mathbb{R}_{\ge0}^M$ (one weight per dimension). We approximate the posterior over $\mathbf{w}$ by an ensemble of $N$ weighted particles $\{(\mathbf{w}^{(i)},\pi^{(i)})\}$ ($N{=}200$ by default). 
% The prior factorizes as $\prod_j \text{Exp}(w_j;\lambda_j)$ - the maximum-entropy choice on $\mathbb{R}_{\ge0}$ given a mean, with finite density at zero so a particle is never penalized for assigning negligible weight to an irrelevant axis. 
Weights are normalized only during utility computation, so adding a new axis mid-dialogue via discovery, 
% which can introduce dimensions absent from any fixed schema) 
never dilutes existing dimension weights. Preference signs are encoded in the loadings $\phi_d$, so non-negative weights retain full expressiveness. The prior and standard degeneracy-control procedure for maintaining particle diversity are in Supplementals sections A.2 and A.3.

\textbf{Information state.} Each particle induces a ranking $\sigma^{(i)}$ of the alternatives by descending utility. The ranking entropy $H(\text{ranking}) = -\sum_\sigma \hat P(\sigma)\log\hat P(\sigma)$, with $\hat P(\sigma)=\sum_{i:\sigma^{(i)}=\sigma}\pi^{(i)}$, summarizes  
the concentration of the posterior and provides a core convergence metric.
% how concentrated the belief is on a single recommendation and is the system's central convergence statistic.
\looseness=-1

\textbf{EIG overview}. GUIDE scores candidate questions by their expected reduction in ranking uncertainty, computed by counterfactually updating the posterior for each possible outcome. It evaluates candidates across all question types on this common information-gain scale and selects the highest-scoring question (details in Supplementals section C.1).

\textbf{Adaptive selection with heterogeneous question types.} Extending EIG across heterogeneous question types introduces a challenge: greedy maximization can favor types with larger nominal one-step gains, even when those gains are rarely realized, while highly informative answers can concentrate the posterior and impair subsequent inference. We therefore augment raw EIG with adaptive corrections for particle degeneracy, recent question effectiveness to reduce uncertainty and exploration, while deprioritizing resolved preference dimensions (details in Supplementals section C.2).

\textbf{Symbolic preference knowledge.} GUIDE complements its numeric preference posterior with a symbolic layer for categorical and conditional preferences that cannot be represented as simple axis weights. Rules may be hard, with violating alternatives being disqualified, or soft, with violations penalized according to the user’s confidence in the rule and the degree to which the alternative violates it. They come from either population-level rules mined offline from survey data or user rules elicited during dialogue (details in Supplementals section D.3).

\textbf{Domain Knowledge Inputs.} The machinery above assumes two things are given before a dialogue starts: the axis set $\mathcal{D}$ with its loading mappings, and a prior over the weights $\mathbf{w}$. 
% alr mentioned above
% OPEN obtains its features by prompting an LLM to describe a domain, so the resulting utility parameterization is only as predictive as the model's confabulations, and a fresh user starts from an uninformative prior. 
GUIDE derives both from a domain-specific population dataset: a table of respondent-level records whose columns are survey responses and attributes. The axes are therefore grounded in measured data about real decision-makers, and a new user's prior is 
seeded based on
% warm-started from 
the population segment they match, avoiding cold-start. This requires a small set of developer-defined inputs: behavioral variables that capture relevant user traits, demographic attributes used to identify and categorize users, item-ownership indicators that provide behavioral evidence, and mappings from preference dimensions to the variables that proxy them (definitions and examples in Supplementals Section D.1). These inputs define how GUIDE interprets population data and matches new users to relevant segments, described within the System Architecture.

\section{System Architecture}

\begin{figure}
    \centering
    \includegraphics[width=0.8\linewidth]{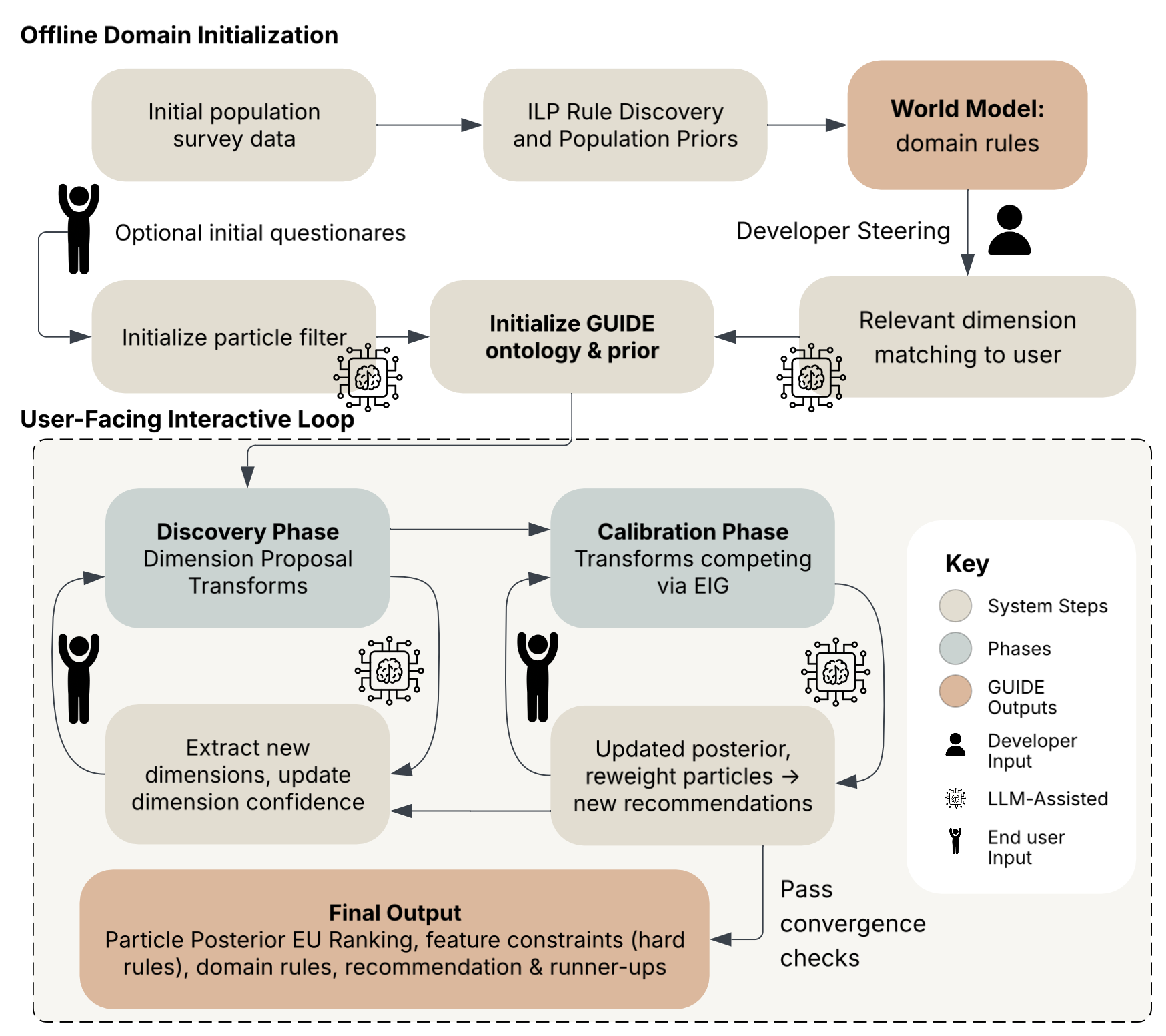}
    \caption{GUIDE Workflow}
    \label{fig:workflow}
\end{figure}

% An overview of the system architecture can be seen in Figure \ref{fig:workflow}. 
Figure~\ref{fig:workflow} is an overview of the system architecture. GUIDE runs an adaptive loop: it maintains a structured posterior over the user's utility dimensions, updates it after each elicitation turn, and selects the next question through a bidirectional two-phase controller that draws candidates from families of question types (which we call ``transforms''). It halts and recommends when the particle posterior has converged.

\subsection{Structured Posterior State}

%Preference vs. Belief: When we are referring to the system's belief about user preferences, I think we can be clearer by calling it a posterior.

The system's representation of the user has three parts: the numeric posterior over utility weights, a symbolic rule set, and an immutable interaction log.

\begin{enumerate}
    \item \textbf{Weight posterior.} The particle ensemble $\{(\mathbf{w}^{(i)},\pi^{(i)})\}$, which carries all graded information about how much each axis matters.

    \item \textbf{Dimensions $\mathcal{D}$} - the active axes, each with a loading mapping $\phi_d$ and a Beta confidence $\text{Beta}(\alpha_d,\beta_d)$. This confidence gates proposed axis admission, shelves underperforming axes, and scales an axis's influence on the posterior by tilting particle log-weights toward particles that place weight on well-supported axes.
    \item \textbf{Preference rules $\mathcal{R}$} - the hard and soft constraints and conditionals currently believed to apply to this user over dimension weights or feature thresholds. Rules come from two sources: population rules injected at calibration and user rules elicited during dialogue. Each rule has a confidence reflecting how strongly it applies to the user; low confidence weakens a soft rule rather than disabling it. The two confidences answer different questions: Beta confidence asks whether an axis is worth modeling, while rule confidence asks how much to trust a constraint.

\end{enumerate}
The interaction log is the raw record of every response, replayed whenever the particle set is rejuvenated so that resampling never loses evidence. Regarding rules, soft rules shape posterior inference by down-weighting violations, while hard rules enforce categorical constraints during recommendation (details in Supplementals Section A.4)

\subsection{Two-phase Question Selection}

\textbf{Phase 1: Discovery.} Only dimension proposal is active; all other transform types (see next section) are suppressed until the basis of relevant axes settles. Discovery runs for at least two elicitation turns and up to n (user-defined) turns, until the best proposal's EIG becomes negligible or plateaus.

\textbf{Phase 2: Calibration.} Pairwise, rule, and dimension-dominance transforms compete via EIG to refine weights on the discovered dimensions. On each cycle, the winning transform asks its question and updates the posterior; a background pass proposes new dimensions, after which posterior degeneracy and convergence are checked.
Movement between discovery and calibration is bidirectional, but the back-edge fires only when the background pass discovers a new dimension. GUIDE injects population priors at the transition to calibration, personalizing preference inference and the recommendation action space without biasing initial exploration (details in Supplementary Section A.5).

\subsection{Elicitation Question Transforms}
Each transform exposes three operations: it maps a response to a likelihood, proposes a ranked list of candidate questions with their EIG, and returns its single best (highest-EIG) candidate. On every cycle, all competing transforms produce their candidates as lightweight templated descriptions, and only the one candidate that wins the cross-transform EIG competition is then rendered into a natural-language question. This bounds the LLM cost of question selection. GUIDE currently supports the
following transforms:

\textbf{Pairwise comparison} A forced choice between two alternatives, with a Bradley-Terry likelihood on their utilities, $P(a\succ b) \propto \exp(u_a/T)$, temperature $T{=}0.5$. Candidates are pairs drawn from the catalog and perturbations, scored by EIG. Greedy selection and an optional lookahead improves question selection by considering whether an initial question enables a more informative follow-up rather than optimizing each question individually, enabling higher-information elicitation sequences (details in Supplementals Section B.1)

\textbf{Rule Elicitation.} This transform builds the symbolic layer, capturing categorical and conditional preferences that weight updates cannot express. An answer updates the posterior; a confirmed rule is added to $\mathcal{R}$ with its elicited hardness and confidence. Candidates come from two signals: \textit{contradiction detection} identifies highly ranked alternatives that violate a rule the user has already stated, while \textit{threshold and dependency probing} surfaces candidate feature cutoffs, weight bounds, and conditionals from patterns in the alternatives. Candidate rules are then filtered to avoid probing for constraints that are unlikely to exist and scored by EIG like other candidates (details in Supplementals section B.2)

\textbf{Dimension dominance ($d_a\succ d_b$).} Direct comparisons of preference dimensions use a Bradley–Terry likelihood to update the log-ratio of weights, shifting the relative importance of dimensions without changing their magnitudes.

\textbf{Free-text statement} A distinctive capability of GUIDE is turning unstructured, free-text preference statements directly into quasi-likelihoods on the same particle posterior used by structured questions.
% -- to our knowledge a novel form of parametric Bayesian inference driven by natural language, rather than the usual pipeline that first parses text into discrete slots. 
To our knowledge, this is a novel form of parametric Bayesian inference driven by natural language. This transformation is triggered by any free-text turn, including the opening message and open-ended or contrastive clarifications. A single LLM call scores every active dimension, returning a signed activation per axis whose polarity is computed relative to that axis's $\phi_d$ (e.g., "I love volatility" indicates higher risk tolerance). The activation vector is normalized and mapped to a bounded, prior-centered log-likelihood update, with its L2 norm determining evidence strength similarly to structured transforms, so text and forced choices accumulate in one posterior. The call also returns hints for concepts with no matching axis, which seed dimension proposal.

\textbf{Dimension proposal.} This transform grows the axis set rather than refining an existing one, running throughout discovery and as a background pass during calibration. Admission is structural: a candidate axis must be grounded in a non-empty feature mapping, cover a feature or direction not already mapped, and remain internally coherent (no anti-correlated features). EIG ranks which existing axis to probe further, since well-established axes yield little clarification value. Accepted axes are seeded with a prior scale borrowed from similar existing dimensions; rejected candidates lower the axis's Beta confidence for later reconsideration (details in Supplementals Section B.3).

A utility update runs for every transform. Each cycle reconciles the posterior with the current state (details in Supplementals section B.4). GUIDE's LLM-dependent steps described in transformations above run on Claude Opus 4.7.

\subsubsection{Decision Alternatives and Perturbations}
Since decision alternatives serve as both question material and recommendation candidates, users whose optima fall between two entries pose a limitation. GUIDE therefore adds structured perturbations around a baseline entry to the candidate set at calibration entry, allowing recommendations to reach interior points. We disable this mechanism during evaluation so all methods share the same discrete action space.

\subsection{Convergence, verification, and recommendation}

\textbf{Stopping.} Convergence fires when ranking entropy remains low and stable for several consecutive cycles, after a minimum number of cycles have elapsed.
% Convergence fires when, for several consecutive cycles, ranking entropy 
% is low (mass concentrated on one alternative), its trend is flat, 
% remains low and stable, and a minimum number of cycles have elapsed. 
Adversarial checks also confirm the recommendation (Supplementals section E.1) 

% \textbf{Adversarial verification.} Before finalizing, a few adversarial pairwise probes challenge the leading recommendation, each chosen to maximize flip fragility. The recommendation is confirmed only if it survives unchanged; otherwise, the system returns to calibration with the new evidence.

\textbf{Recommendation.} GUIDE considers the action space catalog along with calibration perturbations, allowing it to recommend interior points between published alternatives. The candidate set can be adaptively re-centered when user choices favor a perturbation over the current baseline, enabling the search to follow the user's preferred region of the action space. At recommendation, GUIDE selects the alternative with the highest expected utility under the converged posterior: soft rules shape posterior preferences during inference, while hard rules filter out infeasible alternatives. GUIDE also returns the runner-up, feature differences from the baseline, and a confidence measure based on the utility gap between the top two alternatives. Full candidate-generation, re-centering, and scoring details are in Supplementals section E.2.

\subsection{Offline Domain Initialization}
An offline pipeline transforms a population dataset into a set of rules that soft-initialize a Bayesian prior for the domain. This is our domain-initialization contribution. In our portfolio domain, we use the National Financial Capability Study (NFCS) Investor Survey, which provides population-representative investor characteristics over portfolio attributes including risk, volatility, and expected return \cite{finra2024_investor_survey}. Our pipeline produces \textbf{three distinct artifacts} (more details in Supplementals D.4):

% made into citation instead of footnote
% \footnote{https://finrafoundation.org/knowledge-we-gain-share/nfcs/data-and-downloads} 

\begin{itemize}
    \item \textbf{World model} - Background knowledge linking observable signals to latent preference dimensions and their directions, drawn from developer-declared proxies and empirical co-occurrences mined from the data (e.g., "crypto owners tend toward high risk tolerance"). Grounds which dimensions are meaningful for the domain. 
    \item \textbf{Dimension parameterization} - The axis set $\mathcal{D}$, each dimension bound to a signed loading mapping $\phi_d$ over the alternatives' feature space (e.g., "risk tolerance" loads negatively on risk level and volatility). This defines the particle-filter axes. 
    \item \textbf{Population prior} - A per-dimension distributional prior from population domain data, warm started from a user's intake profile to avoid cold-start. 
\end{itemize}

\subsubsection{Segmentation and Discriminative Rule Discovery}

We use Inductive Learning of Answer Set Programs (ILASP)~\cite{law2020ilaspinductivelearninganswer} to mine rules from an input population dataset: each row's demographic and item-ownership values form a candidate antecedent, and its binned behavioral-proxy value the consequent. Segments deviating from the population supply positive examples, with the remainder as negatives. A rule is emitted for a (segment, dimension) pair only when the segment's mean proxy score deviates from the population mean beyond a threshold, with confidence given by the fraction of segment members on the asserted side. The mined rules and accompanying statistics populate the three artifacts above, and pass through the developer steering checkpoint before influencing any live session. ILASP definition, binning methodology, deviation-threshold formulas, and examples are in Supplementals section D.2.

\subsubsection{Developer Steering Checkpoint}

The mining stage outputs candidate population preference rules: statements that a population segment (ex: respondents in an older age band or holders of a particular asset type) tilts a dimension's prior in a given direction. Statistical validity does not necessarily imply a rule is plausible within a domain, so no mined rule reaches a live session without developer review. The architecture makes this checkable by design: axes are declared symbols, and mined rules are symbolic statements over them with explicit confidences and prior strengths. Every quantity that shifts a particle prior is therefore individually inspectable and editable. This is the concrete sense in which GUIDE is steerable. In contrast, an LLM-derived feature ontology is not: the domain-to-feature mapping is latent in the model's weights and can be neither audited nor corrected.

A review checkpoint therefore sits between rule discovery and prior injection. Each rule is rendered as one English sentence by an LLM call, and the domain developer may accept, drop, or edit it, with every conjunct (dimension, direction, threshold, confidence, prior strength) editable and validated before it is cached for reuse. Since the effect of each edit on the prior is explicit, the developer can adjust model weights transparently rather than through LLM prompting (full checkpoint mechanics in Supplementals Section D.5).

\section{Evaluation Methodology}

We evaluate GUIDE's ability to efficiently and accurately recover known preferences in persona-agent simulations conducted in the domain of portfolio investment.
% We numerically evaluate GUIDE against 2 LLM baselines each restricted to 2 question types, 3 ablated versions of GUIDE, and 2 prior work PE methods for a total of 10 methods (including GUIDE). The evaluation task is to answer which elicitation strategy recovers the client's true preferences most efficiently and accurately when given a fixed action space of potential model portfolios and varied investor client personas with different investment goals and strategies.
We compare GUIDE against 4 LLM baselines, 3 ablated GUIDE versions, and 2 state-of-the-art PE methods from prior work (\citealp{handa2024bayesianpreferenceelicitationlanguage}, \citealp{Austin_2024}). 
The evaluation task is to minimize regret in portfolio recommendations in the fewest possible question-answer cycles, which we test using a fixed action space of potential model portfolios and varied client personas representing investors with different strategies.
This enables us to evaluate these nascent systems against a ground truth while accounting for plausible user heterogeneity. 
% This is not intended to substitute for evaluation with human subjects.

\subsection{Comparison Methods}
% We use the following comparison methods:

\subsubsection{LLM Baselines}
We evaluate against two frontier LLMs: OpenAI's GPT-5-4 mini and Anthropic's Claude Opus 4.7. For each, we run conditions where the LLM is restricted to ask either open-ended text or pairwise comparisons, following prior work on LLM PE (e.g., \citealp{li2023elicitinghumanpreferenceslanguage}, \citealp{He_2023}, \citealp{choudhury2026bedllmintelligentinformationgathering}, \citealp{liu2024dellmadecisionmakinguncertainty}).
This yields four LLM baselines (GPT Open-Ended, GPT Pairwise, Claude Open-Ended, Claude Pairwise), representing out-of-the-box LLM approaches available to potential GUIDE adopters.
% Following the open-ended unstructured text and pairwise elicitation strategies as standalone baselines, we implement two question-asking strategies, each run with 2 types of latest frontier LLMs: OpenAI's GPT-5-4-mini and Anthropic's Claude Opus 4.7. There are 4 resulting LLM baselines: GPT Open-Ended, GPT Pairwise, Claude Open-Ended, and Claude Pairwise.

\subsubsection{GUIDE Ablations}
We evaluate the contributions of three key components of GUIDE: (i) domain-specific prior setting, (ii) an elicitation phase for dimension discovery, and (iii) diverse preference state transforms beyond canonical pairwise comparisons. We implement telescoping ablations introducing these system features sequentially, allowing us to assess how each contributes to elicitation quality.
% A telescoping ablation adds population priors, dimension discovery, and representational transforms in sequence. The weakest variant uses flat priors (no domain initialization), a fixed ontology (no discovery), and pairwise questions only, while the strongest variant (strongest ablated variant) adds back population priors \& domain knowledge, and discovery, while still restricting elicitation to pairwise questions. 
Table \ref{tab:guide_variants} describes these ablated variants. "DI" (domain initialization) refers to population priors and domain rule mining.

\begin{table}[t]
\centering
\caption{GUIDE ablation variants from weakest to strongest.}
\label{tab:guide_variants}
\small
\begin{tabular}{p{3.6cm}ccp{1.3cm}}
\toprule
\textbf{Label} & \textbf{DI} & \textbf{Discovery} & \textbf{Ques.Types} \\
% \textbf{Label} & \textbf{DI} & \textbf{Discovery} & \thead{Ques.\\Types} \\
\midrule

No priors, no disc, pairwise &
No &
No &
Pairwise \\
\midrule

Priors, no disc, pairwise &
Yes &
No &
Pairwise \\
\midrule

Priors, disc, pairwise &
Yes &
Yes &
Pairwise \\
\midrule

GUIDE Full &
Yes &
Yes &
All \\
\bottomrule

\end{tabular}
\end{table}

\subsubsection{PE Methods from Prior Work}
We compare to two recent state-of-the-art methods from prior work: OPEN~\cite{handa2024bayesianpreferenceelicitationlanguage} and PEBOL~\cite{Austin_2024}. 
The comparison between the most ablated version of GUIDE and OPEN tests the value of using empirically grounded vs LLM-generated preference dimensions. Comparisons between other methods and PEBOL test the value of each method's parameterization of preference dimensions versus PEBOL's approach of learning a decision boundary directly on the alternatives.

% With access to the codebase, we

We implement OPEN's OEDModel 
% (~\citet{handa2024bayesianpreferenceelicitationlanguage}) 
and PEBOL's PE algorithm (PE with Bayesian Optimization augmented LLMs),
% ~\cite{Austin_2024}
using GPT-5.4-mini for each method's LLM-dependent step (ontology generation and query verbalization for OPEN; dimension identification and question generation for PEBOL). Implementation details in Supplementals section F.1.

\subsection{Experimental Setup}
\begin{figure*}
    \centering
    \includegraphics[width=0.9\linewidth]{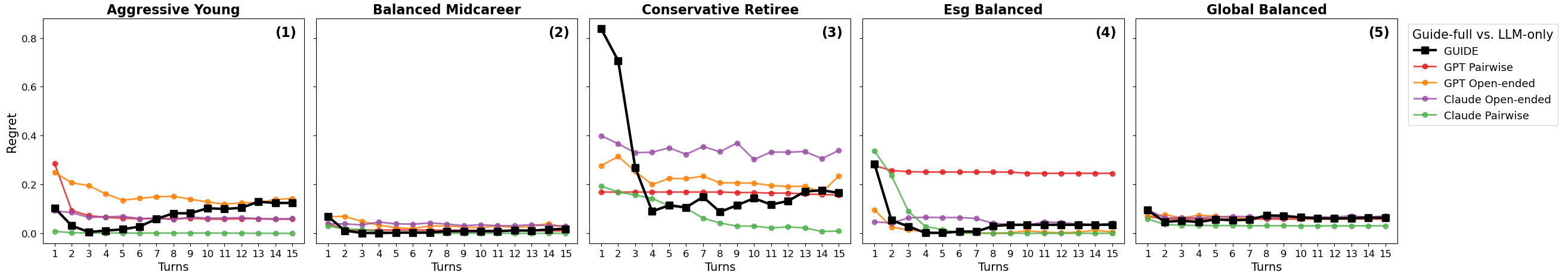}
    \caption{
    % \textbf{Time-Integrated Regret Metrics} 
    \textbf{GUIDE vs. LLMs regret} over elicitation turns
    for investor personas averaged across 50 simulations}
    \label{fig:regret_chart_llm}
% \end{figure*}

% \begin{figure*}
    \centering
    \includegraphics[width=0.9\linewidth]{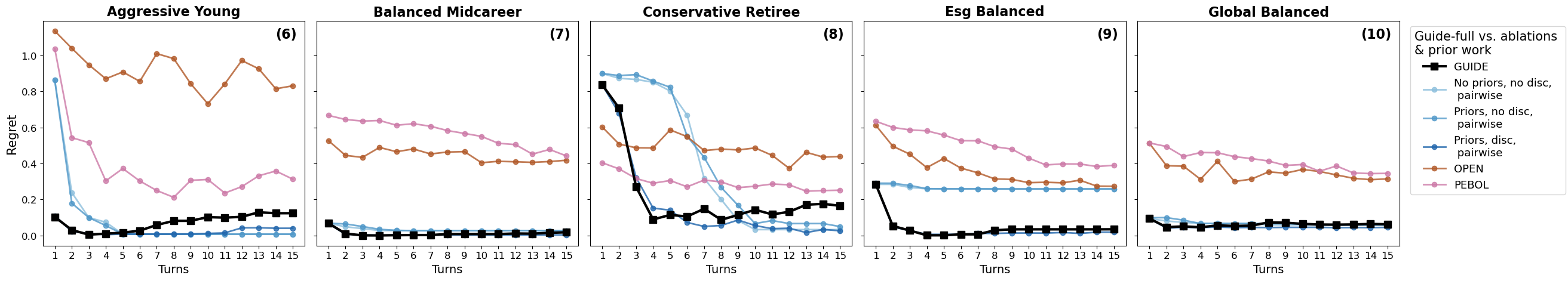}
    \caption{
    % \textbf{Time-Integrated Regret Metrics} 
    \textbf{GUIDE vs. ablations \& prior work regret} over elicitation turns
    for investor personas averaged across 50 simulations}
    \label{fig:regret_chart_work}
\end{figure*}

% We aim to answer how efficiently preference-elicitation methods identify an investor's utility-maximizing portfolio.
We measure how efficiently different PE methods identify the utility-maximizing portfolio for different hypothetical investors. All methods choose from the same fixed catalog of 75 portfolios, generated using an LLM from publicly available model-portfolio data from the Fidelity and Vanguard investment websites. We disable GUIDE's continuous portfolio perturbations so all methods use the same discrete action space. Our comparison uses 50 independent simulations, with 15 elicitation turns per persona. At each turn, every method recommends one portfolio from the same catalog. 

Each persona agent includes a description, opening message, and hidden linear utility function over portfolio features; its latent preference weights determine ground-truth preferences and simulate client choices but are never exposed to the elicitation methods (details in Supplementals section F.2). Cycle 1 measures the regret of the recommendation produced after the elicitation method receives the persona's initial message and completes one elicitation turn.

\subsubsection{Evaluation Personas}
We evaluate five persona agents spanning investment strategies of varying complexity. Three \textit{baseline} personas represent canonical investor profiles spanning distinct, interpretable regions of the investment life cycle and levels of risk tolerance~\cite{NBERw3954}: 1) Aggressive Young (long-term growth, high risk tolerance), 2) Balanced Mid-career (diversified across growth, income, risk, fees, and tax), 3) Conservative Retiree (capital preservation, income stability, and liquidity with explicit risk/volatility limits). Two diverse profiles add more nuanced, multi-dimensional evaluation cases: 1) ESG-balanced (sustainability, financial performance, and portfolio balance), 2) Global-balanced (values geographic breadth, concentration, and volatility). These \textit{diverse} profiles test whether an elicitation method can discover less salient, domain-specific dimensions and identify trade-offs among competing objectives.

All elicitation methods (including GUIDE) interact with the same simulated client LLM, Claude Sonnet 5. The agent converts each persona’s latent weights and constraints into qualitative preferences and responds naturally without directly revealing the underlying dimensions or weights. 
% Questionnaire context is removed so GUIDE receives no intake-derived information unavailable to the baselines.

\subsubsection{Evaluation Metrics}

The primary metric is regret relative to each investor persona's ground-truth utility function $u_p(\cdot)$. For persona $p$, simulation run $s$, and model recommendation $x_{p,t}^{(s)}$ at turn $t$, regret against the catalog-optimal recommendation $x_p^*$ is $R_{p,t}^{(s)}$. We average over $S{=}50$ simulation runs to obtain a mean regret curve per persona and method over $T{=}15$ elicitation cycles to obtain $\bar R_{p,t}$:
\vspace{-6pt}
\[
R_{p,t}^{(s)}=u_p(x_p^*)-u_p\!\left(x_{p,t}^{(s)}\right), \quad
\bar R_{p,t}=\frac{1}{S}\sum_{s=1}^{S} R_{p,t}^{(s)},
\]
\vspace{-6pt}

Lower regret is better (where zero is an optimal recommendation). To measure convergence efficiency independently of a method's starting regret, we report the \emph{normalized regret reduction} (NRR) rate, computed on the mean curve $\bar R_{p,t}$:
\[
\text{NRR}_p=\frac{\bar R_{p,1}-\bar R_{p,t_{\min}}}{t_{\min}\cdot \bar R_{p,1}},
\qquad
t_{\min}=\operatorname*{arg\,min}_{t}\ \bar R_{p,t},
\]
where $\bar R_{p,1}$ is the mean regret at the first turn and $\bar R_{p,t_{\min}}$ is the minimum mean regret, first attained at turn $t_{\min}$. NRR captures the fraction of initial regret eliminated per turn up to the method's best performance, while normalizing for differences in cold-start regret that could otherwise make methods with worse initial performance appear to converge faster.

\section{Evaluation Results}
We report 3 main findings from our evaluation that provides evidence for our 3 initial requirements: (i) \textit{expressive querying}~(R1) allows for efficient use of elicitation cycles to converge and reach minimal regret faster and more consistently than prior work and ablated versions (ii) \textit{empirical grounding}~(R2) minimizes cold start, as GUIDE starts at substantially lower regret than prior work and ablated versions by grounding early interaction in symbolic domain initialization; and (iii) \textit{observability}~(R3) within GUIDE’s architecture exposes interpretable signals of dimension coverage and parameter recovery, creating opportunities for developer oversight, model steering, and avenues of PE research.

\begin{table}[t]
\centering
\footnotesize
\setlength{\tabcolsep}{1.6pt}
\caption{NRR averaged across simulations per persona. Higher value means greater efficiency in regret eliminated per turn (highest bolded). (Disc. is discovery, Agg. is aggressive, Bal. is balanced, Cons. is conservative, PW is pairwise)}
\label{tab:efficiency}
\begin{tabular}{lcccccc}
\toprule
& \multicolumn{2}{c}{Prior Work} & \multicolumn{4}{c}{GUIDE Ablations} \\
\cmidrule(lr){2-3} \cmidrule(lr){4-7}
Persona
& OPEN
& \makecell{PE-\\BOL}
& \makecell{No priors\\No disc.\\ PW}
& \makecell{Priors\\No disc.\\ PW}
& \makecell{Priors\\Disc.\\ PW}
& GUIDE \\
\midrule
Agg. Young      & 0.036 & 0.099 & 0.198 & 0.198 & 0.237 & \textbf{0.316} \\
Bal. Mid.   & 0.023 & 0.022 & 0.100 & 0.100 & 0.245 & \textbf{0.246} \\
Cons. Ret.  & 0.032 & 0.030 & 0.096 & 0.063 & 0.075 & \textbf{0.112} \\
ESG Bal.          & 0.037 & 0.028 & 0.011 & 0.014 & 0.244 & \textbf{0.248} \\
Global Bal.       & 0.069 & 0.024 & 0.038 & 0.034 & \textbf{0.281} & 0.128 \\
\bottomrule
\end{tabular}
\end{table}

\subsection{R1: Expressive Querying for Efficient \& Consistent Convergence}

GUIDE makes efficient use of a limited interaction budget. In Figure~\ref{fig:regret_chart_llm}, GUIDE's regret in early elicitation turns reaches a minimum as quickly as or faster than LLM approaches. For the Aggressive Young and Conservative Retiree personas (panels 1 and 3), GUIDE achieves minimum regret faster; remaining personas also show consistent, competitive performance.
In Table~\ref{tab:efficiency}, GUIDE's NRR exceeds that of both prior-work systems on all five personas and that of the strongest ablation on four of five, 
% In Table~\ref{tab:efficiency}, GUIDE's NRR is higher than that of ablated and prior-work PE systems for 4 out of 5 personas, with NRRs of 0.316 and 0.248 for the Aggressive Young and ESG-Balanced personas, respectively, 
demonstrating greater efficiency and faster convergence. Although prior methods also employ Bayesian inference, their restricted elicitation and initialization mechanisms result in slower convergence.

GUIDE's diverse question types improve early elicitation. In Figure~\ref{fig:regret_chart_work}, GUIDE full consistently matches or outperforms the "Priors, disc, pairwise-only" variant through turn 5. Beyond that, the two diverge, particularly for Aggressive Young and Conservative Retiree (panels 6 and 8): GUIDE reaches a minimum quickly but fluctuates thereafter, whereas the pairwise variant declines more steadily and achieves lower final regret. We attribute this bounce-back to competition among diverse transforms: a late-stage question from a less recently used transform may appear informative under EIG while perturbing an already well-fit posterior. This reveals a trade-off between early efficiency and late-horizon monotonicity: applications prioritizing high-utility recommendations with costly or limited interactions should prefer the full system, while longer-horizon settings may favor a simpler transform set. We evaluate all methods for 15 turns to compare behavior under a common interaction budget. In deployment, GUIDE's stopping mechanism halts upon detecting a plateau, and our observations corroborate this stopping rule.

% \looseness=-1
\subsection{R2: Empirical Grounding for Lower Regret}

GUIDE achieves substantially lower regret during the early stages of elicitation. Across 50 simulations and all personas, we aggregate regret values into a distribution at each elicitation turn and report its mean (table in Supplementals section G.1). GUIDE begins with a mean regret of $0.278$ at turn 1, compared with $0.678$ for OPEN and $0.651$ for PEBOL. By turn 5, GUIDE's regret falls to $0.038$, which is over $90\%$ lower than both OPEN and PEBOL's regret at the same turn.
% $93.2\%$ lower than OPEN's regret and $91.8\%$ lower than PEBOL's regret at the same turn. 

GUIDE's early advantage is from structural learning before elicitation begins: symbolic domain initialization provides population-grounded priors, while dimension discovery identifies relevant preference axes during interaction. The GUIDE ablation retaining both components but restricted to pairwise comparisons achieves similarly low initial regret ($0.279$) in the same averaged results, whereas regret in variants without initialization/discovery exceeds $0.44$. This pattern also holds across individual investor personas (Figure~\ref{fig:regret_chart_work}). These results suggest that grounding the initial preference representation in population-level knowledge and refining it through dimension discovery reduces cold-start effects for faster convergence to high-quality recommendations.

\begin{figure}
    \centering
    \includegraphics[width=\linewidth]{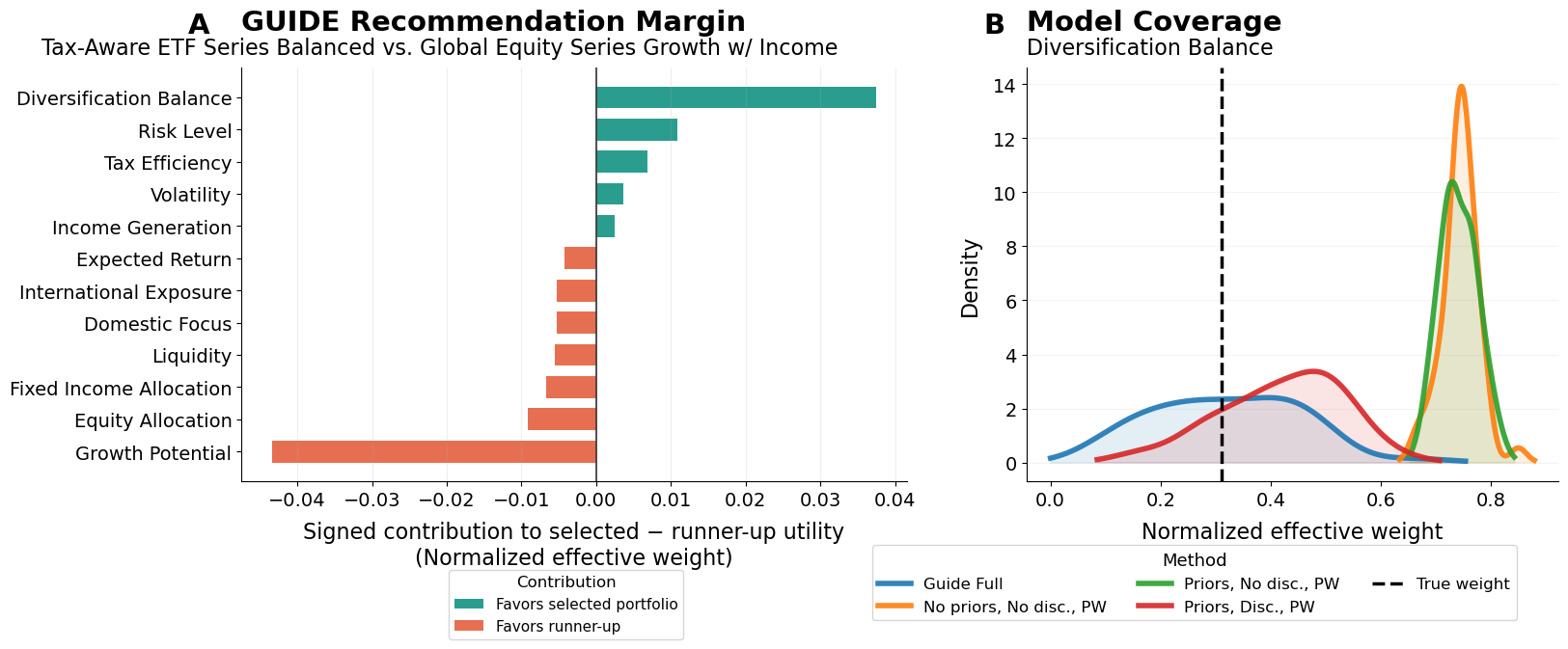}
    \caption{\textbf{GUIDE observability} for Bal. Midcareer simulation: (A) dimension contributions to recommendation margin, (B) diversification balance dimension model coverage}
    \label{fig:observ_2}
\end{figure}

\subsection{R3: Observability into Model Coverage and PE}
GUIDE provides observability into the PE process through its explicit posterior over latent preference dimensions, tracked across elicitation turns. Developers can use this to understand recommendation selections, diagnose unstable or disagreeable outputs, and identify dimensions that may be missing or misclassified during domain initialization or steering.

Figure~\ref{fig:observ_2} illustrates these capabilities for the Balanced Midcareer persona. Panel A shows the recommendation's margin over the runner-up for each dimension in a simulation run, capturing the difference between the recommended and second-best weights. This enables post-hoc analysis of which preferences drive a recommendation selection over suboptimal ones. This view is particularly helpful when outputs become unstable or disagreeable, as developers can examine how GUIDE's posterior influences recommendations across elicitation cycles. 
% Developers can inspect posterior trajectories to diagnose inference failures caused by incorrect domain initialization or steering that leads to dimensions being persistently overweighted or drifting over question-and-answer turns. 
Inspecting these posterior trajectories can help diagnose inference failures arising from incorrect domain initialization or steering, such as dimensions that remain overweighted or drift across elicitation turns. Model coverage analysis can diagnose such failures: during development, in-silico experiments can compare GUIDE's inferred posterior with the hidden weights of a simulated persona. Panel B, for example, shows the distribution of weights for a specific dimension (e.g., diversification balance) across 50 simulations. Both full and ``priors, discovery, pairwise" ablations indicate weight distributions overlapping with ground truth, and can be fine-tuned further to increase accuracy. In contrast, the remaining ablations, which omit both domain initialization and discovery, produce incorrectly concentrated weight distributions. Such analyses can help identify gaps or mis-specification in the modeled preference dimensions, informing revisions to initialization rules, vocabulary, or query transformations as a form of sensitivity analysis.

\section{Conclusion and Future Work}

We present GUIDE, a novel LLM-assisted Bayesian PE framework 
% that combines adaptive selection across diverse question types with symbolic domain knowledge 
to initialize domain-specific preference models. 
Our evaluation demonstrates that GUIDE improves elicitation efficiency consistently across early interactions via expressive querying and empirical grounding while observability within our architecture allows analysis of the PE process.

\looseness=-1

Future work should evaluate GUIDE with human subjects in various applications, and our architecture design facilitates this research. GUIDE's structured and observable approach enables comparative evaluation of diverse question types and selection policies influencing elicited preferences. Similarly, GUIDE's affordances for developer steering create opportunities to deploy and study PE systems across domains. 

GUIDE is a domain-general PE framework integrating adaptive learning, symbolic reasoning, and human oversight. Its extensible architecture supports development and evaluation of (1) new elicitation question types and interfaces, (2) corresponding question selection policies, (3) preference uncertainty representations and learning methods, and (4) domain logic. 
% derived from various data sources 
GUIDE's preference state also provides a portable user profile that can support personalization in downstream applications. Ultimately, we envision GUIDE as a step toward PE systems that advance human-AI alignment by combining adaptive learning, symbolic reasoning, and human oversight across diverse decision-making domains.

\bibliography{aaai2027}

\appendix
% s a little breathing room below the title
\section*{Supplemental Materials}
% \section{Bayesian Inference Machinery}
\section{Particle Posterior Mechanics}
% \label{app:prelim}

\subsection{Dimensions and loading mappings.} A preference dimension $d$ is a latent evaluative \textit{axis} (ex: "risk tolerance") bound to the feature space by a signed loading mapping $\phi_d$, which scores how well an alternative satisfies that axis. Specifying an axis consists of giving one signed loading $\lambda_{d,k}\in\mathbb{R}$ per feature, where the sign is the direction in which that feature moves the axis, and the magnitude is its strength as an indicator. The axis score is then the loading-weighted, scale-normalized average:
$$\phi_d(\mathbf{f}) = \frac{\sum_k \lambda_{d,k}\, f_k}{\sum_k |\lambda_{d,k}|}.$$

For example, risk tolerance loads on \{risk level: -1, volatility: -0.9, equity allocation: -0.6, fixed income: 0.6, liquidity: 0.3\}, so a higher weight on this axis favors lower-risk portfolios; binary $\pm1$ loadings are the equal-weight special case. The loadings for the starting axes are declared offline by the developer; axes discovered mid-dialogue are proposed by the LLM in this same format and validated before use. Preferences that peak in the interior of the feature space are expressed by defining a feature that already encodes that shape, rather than by complicating $\phi_d$: for example, a "balanced" investor is captured by a diversification-balance feature that peaks at a 50/50 equity/bond split, on which the balance axis then loads positively. At runtime, we calculate each $\phi_d$ with a small additive model (one shape function per feature, fit by round-robin gradient boosting over the catalog) trained to reproduce the expression above.

\subsection{Prior form.} The prior over the weight vector $\mathbf{w}$ factorizes as $\prod_j \text{Exp}(w_j;\lambda_j)$ -- the maximum-entropy choice on $\mathbb{R}_{\ge0}$ given a mean, with finite density at zero so a particle is never penalized for assigning negligible weight to an irrelevant axis.

\subsection{Degeneracy control and rejuvenation}
Repeated reweighting concentrates the mass on a few particles, so degeneracy is tracked by the normalized effective sample size $\widetilde{\text{ESS}}$. When $\widetilde{\text{ESS}}$ falls below $0.5$, the ensemble is resampled and then \emph{rejuvenated}: each particle's log-weights are randomly jittered, and each move is accepted in proportion to how well the new weights fit the full posterior
(i.e., a Metropolis--Hastings step~\cite{robert2016metropolishastingsalgorithm}).
% - a Metropolis--Hastings step \cite{robert2016metropolishastingsalgorithm}. 
This restores particle diversity without biasing the posterior it targets, and recurs whenever the posterior collapses.

\subsection{Rules affecting posterior} 
Rules act on the posterior at two moments. Continuously, a soft rule penalizes each violating particle's log-weight in proportion to its confidence and to the magnitude of the violation, $\ell^{(i)} \mathrel{-}= 2\,c_r \cdot v^{(i)}$, so violating weight vectors become unlikely without becoming impossible. This lets rejuvenation cross low-mass regions and lets later evidence overturn a mistaken rule. At recommendation time, hard rules act instead as feasibility filters that disqualify violating alternatives outright. This distinction between rules allows graded evidence to be assigned to the weights and categorical requirements in the feasible set.

\subsection{Prior injection at calibration entry.} Population priors are deliberately withheld during discovery, since applying them earlier would bias exploration toward population-mean axes before the user's own axes have surfaced. They are injected in the single cycle that transitions discovery to calibration, and they personalize two different things. On the posterior, a domain dimension prior (plus any prior survey data) nudges the particles toward population-relevant axes. GUIDE then also personalizes the action space: the constraint solver applies the population- and client-matched rules and selects a baseline alternative (e.g., the constraint-optimal portfolio for this client), around which a catalog of perturbations is generated. Personalizing the action space, not just the weights, is what allows the eventual recommendation to reach a user's interior optimum instead of being confined to the alternatives the developer and domain happened to publish.

\section{Elicitation Transforms}

\subsection{Pairwise Lookahead}
Greedy selection takes the single top pair; an optional lookahead instead runs a short, depth-bounded beam search (depth 2, beam width 3) over (question $\to$ outcome $\to$ follow-up) chains, discounting a chain's later gains by $\delta{=}0.85<1$. This lets it prefer a first question that sets up a more informative follow-up, recovering higher-EIG chains where the EIG surface is highly curved. Pairs that are projection-equivalent under $\boldsymbol{\phi}$ (identical utilities, hence zero EIG) are pruned without scoring.

\subsection{Rule Elicitation.} This transform builds the symbolic layer, capturing the categorical and conditional preferences that no weight update can express. An answer updates the posterior through the rule penalty described above; a confirmed rule is added to $\mathcal{R}$ with its elicited hardness and confidence. Candidates come from two signals. Contradiction detection finds an alternative that the current posterior ranks highly but which violates a rule the user already stated, and asks whether that rule is truly inviolable or merely a tendency, and where its cutoff actually lies. Threshold and dependency probing propose candidate feature cutoffs, weight bounds, and conditionals surfaced by frequent-pattern discovery over the alternatives. Since asking about a constraint that does not exist wastes a cycle, candidates are filtered by a per-dimension Dirichlet posterior over the three possibilities $\{$hard, soft, no constraint$\}$, maintained from the outcomes of previous rule probes. Because probing every dimension independently would be slow, evidence is also pooled across \emph{similar} dimensions, where the similarity of two axes is the Jaccard overlap of the feature sets they load on,
$$K(d_a,d_b)=\frac{|F_{d_a}\cap F_{d_b}|}{|F_{d_a}\cup F_{d_b}|},$$
with $F_d$ the set of features appearing in $d$'s loading mapping. Observing an outcome adds a full count to the probed dimension's Dirichlet parameters and a fractional count of $K(d_a,d_b)$ to every other dimension, so learning that no hard cap exists on one risk-loaded axis partially transfers to other axes built from the same features, while axes with no features in common are left untouched.
% and pooled across features similar to the target axis. 
A probe is suppressed when this posterior places little mass on a constraint that exists at all; surviving probes are then EIG-scored like any other candidate.

\subsection{Dimension proposal.} 

This transform grows the axis set and is the one place where the system may add a degree of freedom rather than refine one. It is active throughout discovery and runs as a background pass during calibration. Admission is structural rather than information-theoretic: a proposed axis must clear three gates: 1) grounding (a non-empty signed feature mapping, verified against the distribution of alternatives), 2) feature novelty (it covers a feature/direction not already mapped), and 3) coherence (no two of its features are anti-correlated beyond a threshold, which would make its utility contribution non-monotone). The decision to keep growing the basis rests with the phase controller and the Beta confidence. 
 EIG only ranks the questions this transform can ask: clarification questions about an existing axis are EIG-scored directly, so an axis the posterior already understands scores low and clarification is steered toward the least-settled axes. A passing candidate is added with prior-scale imputation, borrowing its exponential prior scale from existing dimensions in proportion to feature-overlap similarity, so a new axis starts at a sensible magnitude. Since weights are normalized only at utility-computation time, this addition does not dilute the weights already assigned to existing axes. A candidate that fails any of the three gates is discarded for that cycle and recorded in the run log. This additionally keeps discovery auditable: a developer can see which concepts the LLM proposed and which gate each one failed. Discarding is not permanent, as discovery re-runs on every subsequent free-text turn and maintains no blocklist, so a concept rejected only because its proposed mapping was malformed or fully redundant can still be admitted later if the LLM proposes it with a valid, novel mapping. An axis's Beta confidence is moved separately from this structural gating, by the user's own responses about that axis, as confirmations raise it and denials lower it.
% Rejected candidates lower the axis's Beta confidence and are kept in a log for later rephrasing.

\subsection{Utility Updates}
Independently of which transform wins, each cycle reconciles the posterior with the current state: every active dimension is given a particle column, dimension confidences are reapplied, and the rule set is re-evaluated. Rules that the evidence has firmed up are promoted from soft to hard, and a hard rule that the user's own choices systematically violate is demoted, so a single mis-elicited constraint cannot dominate the final recommendation.

\section{EIG Implementation Details}

\subsection{Expected information gain overview}
A candidate question $q$ with outcomes $\mathcal{O}_q$ is scored by the expected reduction in ranking entropy,
$$\text{EIG}(q) = H(\text{ranking}) - \sum_{o\in\mathcal{O}_q} P(o)\,H(\text{ranking}\mid o),$$
where $H(\text{ranking}\mid o)$ is computed by counterfactually reweighting the particles in the posterior under outcome $o$ 
% (using that question's likelihood, so scoring and updating are consistent) 
using that question's likelihood,
and $P(o)=\sum_i\pi^{(i)}P(o\mid\mathbf{w}^{(i)},q)$ is the particle-mixture predictive distribution. To choose the next question, every transform enumerates its own candidate questions and computes each candidate's prospective EIG by this counterfactual reweighting; the question actually asked is the single highest-EIG candidate across all transforms. EIG is thus the common currency in which questions of different structures compete on one scale.

\subsection{EIG Augmentation Formulas}
We augment raw EIG with three multipliers--\emph{ESS discount}, \emph{realized-success}, and \emph{curvature} multipliers--along with two escape mechanisms--\emph{force-switch} and \emph{resolved-axis deprioritization}--to favor productive question types and prevent stagnation.

An ESS discount penalizes questions whose informative answers would collapse the particle set, 
$$\text{EIG}_{\text{disc}}(q)=\sum_{o}P(o)\,\Delta H_o\left(\tfrac{\text{ESS}_o}{\text{ESS}_{\text{cur}}}\right)^{\beta},\qquad \beta{=}2,$$
where $\Delta H_o = H(\text{ranking}) - H(\text{ranking}\mid o)$ is the entropy drop under outcome $o$, i.e. the same per-outcome quantity that the EIG expression above sums over. Here $\text{ESS}_{\text{cur}}$ is the current normalized effective sample size and $\text{ESS}_o$ is the normalized effective sample size the ensemble would have after outcome $o$.The ratio is capped at $1$, so the term only penalizes questions. 
A realized-success multiplier then holds each type accountable to its own track record: each type $T$ keeps an exponential moving average of whether it actually delivered entropy reduction, $s_T \leftarrow (1-\alpha)\,s_T + \alpha\,\mathbb{1}[\Delta H > \epsilon]$ ($\alpha{=}0.5$, $\epsilon{=}10^{-3}$, initialized at $1$), and its EIG is scaled by $s_T$.

The third multiplier is a \emph{curvature bonus} $\operatorname{clip}\!\big(1 + 0.5\tanh(2\kappa),\,0.6,\,1.5\big)$, where $\kappa$ is a finite-difference second derivative of the EIG surface at the candidate question. Intuitively, $\kappa$ asks whether a question sits on a rising part of the information landscape or on top of a bump: $\kappa>0$ means EIG is still climbing in the direction this question probes, so the multiplier rises toward $1.5$ and the question is promoted, while $\kappa<0$ means the question is at or past a local peak and offers diminishing returns, so the multiplier falls toward $0.6$. The $\tanh$ and the clip keep a noisy curvature estimate from overwhelming the raw EIG it multiplies, so this term re-ranks candidates that are already close rather than overriding the information measure. Two escape mechanisms then address local optima in the EIG surface: (1) a \emph{force-switch} that suppresses the dominant question type once the windowed mean $\Delta H<0.005$ while one type fills $\ge 75\%$ of the recent window, breaking the feedback loop in which a locally strong type crowds out the others; and (2) \emph{resolved-axis deprioritization}, which down-weights questions targeting a dimension whose posterior weight variance has already fallen below a floor.

% The three escape mechanisms address local optima in the EIG surface: (1) a \emph{curvature bonus} $\operatorname{clip}\!\big(1 + 0.5\tanh(2\kappa),\,0.6,\,1.5\big)$, where $\kappa$ is a finite-difference second derivative of the EIG surface (positive $\kappa$ means the surface is still climbing); (2) a \emph{force-switch} that suppresses the dominant question type once the windowed mean $\Delta H<0.005$ while one type fills $\ge 75\%$ of the recent window, breaking the feedback loop in which a locally strong type crowds out the others; and 

% (3) \emph{resolved-axis deprioritization}, which down-weights questions targeting a dimension whose posterior weight variance has already fallen below a floor.

\section{Domain World Model Initialization}
\subsection{Developer Domain Inputs}

Developers, or experts initializing GUIDE for elicitation usage in their own domains, supply a small, declarative vocabulary:

\begin{itemize}
    \item Behavioral variables: columns that measure a respondent's standing on a latent trait of the user, each with a scale and sign. Example: the NFCS survey's self-reported "risk willingness" and "financial planning horizon" items.

    \item Demographic variables: categorical attributes used to segment the population. These matter because they are also what a new user can be matched on at intake, making them the conditioning side of every mined rule. Example: age group (young, midcareer, senior), income level (low, mid, high).

    \item Item-ownership indicators: binary flags for revealed engagement. These matter because they are behavioral evidence rather than self-report, often carrying preference signals that stated answers miss. Example: owns a brokerage account, owns crypto.

    \item Dimension specifications: the target preference dimensions (or axes) that GUIDE will reason over, declared here by the developer as dimension parameterizations. Each dimension is listed with the behavioral columns that proxy it and the direction of the proxy. Example: "risk tolerance" is proxied by "risk willingness" with $+1$ (higher risk willingness $\Rightarrow$ higher risk tolerance), while "income preference" uses the same column with $-1$ (low risk willingness implies an income tilt).
\end{itemize}

\subsection{Segmentation and discriminative rule discovery}

We use Inductive Learning of Answer Set Programs (ILASP) \cite{law2020ilaspinductivelearninganswer}. An answer set program is a declarative logic program whose solutions ("answer sets") are the assignments satisfying a set of logical rules, and which supports both hard and soft constraints. This representation suits us for two reasons: the segment-conditioned statements we want are naturally first-order rules, and the soft-constraint weights map directly onto the rule confidences GUIDE needs. ILASP learns from \emph{examples}, which in its formalism are partial interpretations: each example names some atoms that must hold and some that must not. The learned program must include at least one answer set that extends every \emph{positive} example, while having no answer sets that extend any of the \emph{negative} examples. Positive examples are therefore the cases a rule must account for, and negative examples the cases it must not wrongly cover. Given such examples, ILASP induces candidate first-order rules and returns a minimal program consistent with them. We translate the program's soft-constraint weights into context-specific prior strengths and rule confidences (not new world knowledge) to address cold-start. The pipeline has four stages: (1) data pre-computation, (2) ILASP task construction, (3) inductive solving, and (4) rule conversion.

ILASP learns rules from examples, which are the respondent records themselves: each row's demographic and item-ownership values form the candidate antecedent, and its binned behavioral-proxy values the consequent. This means a segment that deviates from the population supplies positive examples, and the rest supply negatives \cite{law2018inductivelearninganswerset}.

Each behavioral proxy is normalized to $[0,100]$ and partitioned into low/medium/high segments by population quantiles (tertiles by default). Binning defines a rule's antecedent, and must be discrete because a logic-program antecedent has to be a checkable predicate - \texttt{senior(X)} either holds or not. Therefore, a continuous score cannot appear directly, and discretizing yields groups ("high-risk-willingness investors", "seniors") whose mean on a dimension can be compared against the population mean. Quantiles rather than fixed cutoffs keep segment sizes balanced regardless of the proxy's raw distribution.

For each (segment, dimension) pair the solver computes the segment's mean proxy score and the population mean, and emits a rule only when the segment deviates from the population by at least $\delta_{\min}$ ($|\text{deviation}| \geq \delta_{\min}$, default as 10 points on the $0$–$100$ scale):
$$\text{deviation}(\text{seg}, d) = \text{segment\_mean}(\text{seg}, d) - \text{pop\_mean}(d)$$

This keeps rules informative rather than trivial: only segments that genuinely depart from the baseline carry conditioning value. The rule's threshold is the population mean itself (the segment is asserted to lie above or below the baseline), and its confidence is the fraction of segment members on the asserted side. A minimum support (default 5\% of the population) prevents rules from tiny segments. For example, if investors aged 55+ score 12 points below the population mean on the "time horizon" proxy, with 80\% of the segment below that mean, the pipeline emits "senior $\Rightarrow$ 'time horizon' below baseline" with confidence 0.8. The mined rules and accompanying unconditional statistics populate the three artifacts above and pass through the developer steering checkpoint before influencing any live session.

\subsection{Symbolic preference knowledge rules}

Not all preference information is naturally expressed as a weight on an axis. Statements like "nothing above 40\% risk" or "if fees are high, I need higher returns" are categorical or conditional: they describe the shape of the feasible region rather than a graded trade-off, and a purely numeric utility either satisfies them only on average or must approximate them with extreme weights. GUIDE therefore keeps a symbolic layer alongside the numeric posterior, holding rules of the forms WeightAbove/WeightBelow (bounds on a dimension's weight), FeatureThreshold (bounds on a feature value), and IfThen (conditionals). Each rule is either 1) hard: a constraint asserted to always hold and disqualifies any alternative violating it, or 2) soft: a tendency that penalizes violating particles in proportion to a confidence and to how bad the violation is, without forbidding them. Two distinct provenances feed this layer, and we keep them separate throughout: population rules, mined offline from survey data and injected once as priors, and user rules, elicited during the dialogue from the individual in front of the system. Population rules describe who the user probably resembles; user rules describe what this user has actually said.

\subsection{Offline Domain Initialization Outputs}

The offline pipeline that transforms a population dataset into a set of rules that soft-initialize a Bayesian prior for the domain. This produces three distinct artifacts:

\begin{itemize}
    \item \textbf{World model} - background knowledge stating which observable data signals indicate which latent preference dimension and direction. This comes in two forms: developer-declared proxies ("risk-willingness proxies risk tolerance positively") and empirical co-occurrences mined from the data ("crypto owners tend toward high risk tolerance"). Converted to natural language, it grounds which dimensions are meaningful for the domain.
    \item \textbf{Dimension parameterization} - the axis set $\mathcal{D}$, each dimension bound to a signed loading mapping $\phi_d$ over the alternatives' feature space. For example, "risk tolerance" $\mapsto \{\text{risk\_level}{:}-1, \text{volatility}{:}-1\}$, so a high weight on this axis favors low-risk, low-volatility portfolios. This defines the particle-filter axes.
    \item \textbf{Population prior} - a per-dimension distributional prior from unconditional population statistics. For "risk tolerance", this yields a Gaussian centered on the population's average normalized risk willingness. These statistics do two things when a session starts: they set the scale of each dimension's weight prior, so a new user's particles begin near the population average rather than at an uninformative default, and they determine which dimensions are seeded at all, since a dimension carrying very little prior mass is not worth starting the filter with. The segment rules mined above shift the population-average starting point toward the segments the individual user matches at intake, helping avoid cold-start.
\end{itemize}

\subsection{Developer Steering Checkpoint: Full Mechanics}
A review checkpoint sits between rule discovery and prior injection, mapping $\mathcal{R}_{\text{mined}} \to \mathcal{R}_{\text{approved}}$ with each entry carrying provenance $p\in\{\text{accepted}, \text{edited}, \text{dropped}\}$. Each rule is rendered as one plain-English sentence by an LLM call, and the domain developer may accept, drop, or edit it. Every edit is checked before it is saved: a rule may only refer to a dimension that exists in the declared vocabulary, each of its conditions must point either above or below its threshold, and its threshold, confidence, and prior strength must each fall in $[0,1]$. An edit failing any of these checks is refused with an explanatory message rather than silently accepted, so a developer cannot leave the rule set in a state the live system would misread. Since the effect of each edit on the prior is explicit, the developer is adjusting the model's weights transparently rather than through prompt engineering. 

An additional affordance follows from the symbolic representation. Specifically, the checkpoint can surface the approved set's joint logical implications (meaning which segments would activate which priors and constraints), exposing conflicting or redundant rules that are invisible rule-by-rule before any of them reach a live session.

\section{Recommendation Details}

\subsection{Adversarial verification.} Before finalizing, a few adversarial pairwise probes challenge the leading recommendation, each chosen to maximize flip fragility. The recommendation is confirmed only if it survives unchanged; otherwise, the system returns to calibration with the new evidence.

\subsection{Recommendation and Search}

The candidate set is the catalog together with the perturbations built at calibration entry, which is what allows the system to recommend an interior point when the user's optimum lies between published alternatives. That set is not rebuilt at the end: it is re-scored under the converged posterior and re-centered during calibration only if the user's pairwise choices favored a perturbation over the standing baseline. In this case, the chosen perturbation becomes the new center. The recommendation maximizes expected posterior utility over the alternatives that satisfy every hard rule,
$$\hat k = \arg\max_{k\,:\,\text{feasible}} \ \sum_i \pi^{(i)} u_k(\mathbf{w}^{(i)}),$$
This is where the symbolic layer's categorical half takes effect: soft rules have already shaped the weights through their penalties, while hard rules apply here as a feasibility filter, removing violating alternatives from consideration regardless of their score. Alongside the choice, GUIDE returns the runner-up (second-best) alternative, the feature deltas relative to the baseline, and a confidence derived from the runner-up gap (the difference in expected posterior utility between the best and second-best alternatives, where a larger gap indicates a more decisive choice).

%\vspace{50pt}
\section{Evaluation Methodology}

\subsection{Prior Work Implementation Details}

We implement OPEN using its OEDModel implementation \citet{handa2024bayesianpreferenceelicitationlanguage}. At the start of each simulation, GPT-5.4-mini generates a ranked ontology of 10 broad, non-obvious binary investment features. This ontology is shared across all personas within a simulation but regenerated between simulations. GPT-5.4-mini maps each portfolio into this binary feature space and verbalizes OPEN's highest expected information gain (EIG) pairwise query. The OEDModel then updates its ensemble utility model based on the persona's response.

We use PEBOL's decision-theoretic PE algorithm (PE with Bayesian Optimization augmented LLMs) \cite{Austin_2024}. Each portfolio is treated as an item described by its name, features, and metadata. At each interaction, GPT-5.4-mini identifies a new investment aspect, generates a binary (yes/no) question, and estimates each portfolio's compatibility after observing the response. The response is then used to update PEBOL's Beta posterior over portfolio utilities.

\subsection{Simulated Persona Agents}

Each simulated investor is a persona agent containing a description, opening message, and a hidden linear utility function
$$
u_p(x)=\sum_d w_{p,d}f_d(x),
$$
where $f_d(x)$ is portfolio $x$'s normalized value on dimension $d$ and $w_{p,d}$ is persona $p$'s latent preference weight. Positive and negative weights, respectively, encode preference for higher and lower feature values. These weights determine both the ground-truth optimum and the simulated client's choices, but are never exposed to an elicitation method.

\section{Additional Evaluation Results}

\subsection{Summary Regret Table}
Table \ref{tab:regret_table_extra} shows the regret values summarized across persona agents for each method. This is referenced within our results.
\begin{table}[t]
\centering
\footnotesize
\setlength{\tabcolsep}{1.7pt}
\caption{Regret values averaged across all simulations and personas for each method}
\label{tab:regret_table_extra}
\begin{tabular}{lcccccc}
\toprule
 & \multicolumn{2}{c}{Prior Work} & \multicolumn{4}{c}{GUIDE Ablations} \\
\cmidrule(lr){2-3} \cmidrule(lr){4-7}
Turn & OPEN & PEBOL & 
\makecell{No priors,\\no disc,\\pairwise} & 
\makecell{Priors,\\no disc,\\pairwise} & 
\makecell{Priors,\\disc,\\pairwise} & GUIDE \\
\midrule
1  & 0.678 & 0.651 & 0.442 & 0.444 & 0.279 & \textbf{0.278} \\
3  & 0.541 & 0.499 & 0.269 & 0.281 & 0.082 & \textbf{0.071} \\
5  & 0.560 & 0.461 & 0.233 & 0.237 & 0.042 & \textbf{0.038} \\
8  & 0.519 & 0.400 & 0.111 & 0.126 & \textbf{0.025} & 0.056 \\
10 & 0.457 & 0.392 & 0.077 & 0.085 & \textbf{0.026} & 0.071 \\
15 & 0.455 & 0.348 & 0.077 & 0.081 & \textbf{0.027} & 0.081 \\
\bottomrule \\
\end{tabular}
\end{table}


\end{document}